\documentclass[letterpaper,twocolumn,10pt]{article}
\usepackage{zhanggroup}

\usepackage{amsmath}
\usepackage{amssymb}
\usepackage{xspace}
\usepackage{multirow}
\usepackage{graphicx}
\usepackage{colortbl}
\usepackage[most]{tcolorbox}
\usepackage{algorithm}
\usepackage{algpseudocode}
\usepackage[shortlabels,inline]{enumitem}
\setlist[itemize]{noitemsep,leftmargin=*,topsep=0em}

\newcommand{\mypara}[1]{\noindent{\bf {#1}.} \xspace}
\newcommand{\OSV}{OSV-5M\xspace}
\newcommand{\YFCC}{YFCC4k\xspace}
\newcommand{\IMG}{Im2GPS3k\xspace}
\newcommand{\MPS}{MP-16\xspace}
\newcommand{\MPP}{MP16-Pro\xspace}

\begin{document}

\date{}

\title{\bf Where or Who?\\Photographer Leakage Flips Visual Geolocation Rankings}

\author{
Yukun Jiang\ \ \
Yage Zhang\ \ \
Mingjie Li\ \ \
Yang Zhang\thanks{Corresponding author (yangzhang.almo@outlook.com).}
\\
\\
\textit{CISPA Helmholtz Center for Information Security}
}

\maketitle

\begin{abstract}
Visual geolocation benchmarks measure whether a model can infer where a photo was taken, a capability that powers applications and threatens location privacy.
However, we find that most of them also reward recognizing who took the photo.
Specifically, in these benchmarks, 55.95\% to 98.24\% of test images come from photographers who also appear in the reference data (training and retrieval images).
To measure this photographer leakage, we build photographer-disjoint reference data for OpenStreetView-5M (OSV-5M) and YFCC4k, each with a size-matched random control.
On OSV-5M, separating photographers lowers the country accuracy of three methods spanning generation, retrieval, and classification by 26.62\% to 39.01\% relative to the control, whereas the control itself changes it by at most 1.24\%.
As a result, rankings flip on both benchmarks, as the previously top-ranked methods (PLONK and GeoRanker) fall to fourth and third, respectively, while two zero-shot vision-language models rise to first and second.
Moreover, attribution shows that the analyzed retrieval methods prefer images of the query's own photographer over other photographers' images with the same DINOv2 similarity and distance.
Hence, standard benchmark rankings partly reflect who took a photo rather than where it was taken, which misleads model selection and understates the location privacy risk of zero-shot vision-language models relative to specialized methods.
To support geolocation evaluation without photographer leakage, we release photographer-disjoint benchmark (WoW-PD) splits.\footnote{\url{https://huggingface.co/datasets/TrustAIRLab/WoW-PD}}
\end{abstract}

\begin{figure}[t]
\centering
\includegraphics[width=\columnwidth]{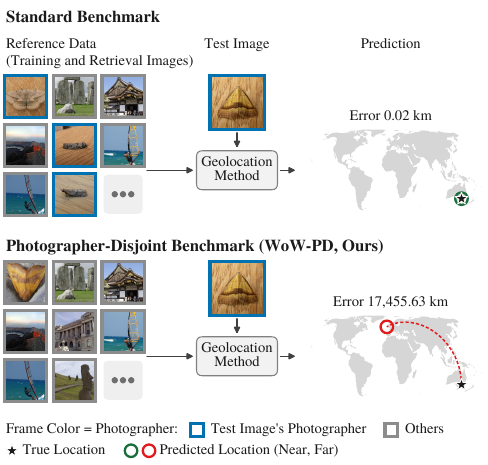}
\caption{Photographer leakage. Nearest-neighbor retrieval locates a \YFCC test image by matching it to other images of its photographer in the reference data, and misses it by thousands of kilometers with photographer-disjoint reference data.}
\label{figure:teaser}
\end{figure}

\begin{figure}[t]
\centering
\includegraphics[width=\columnwidth]{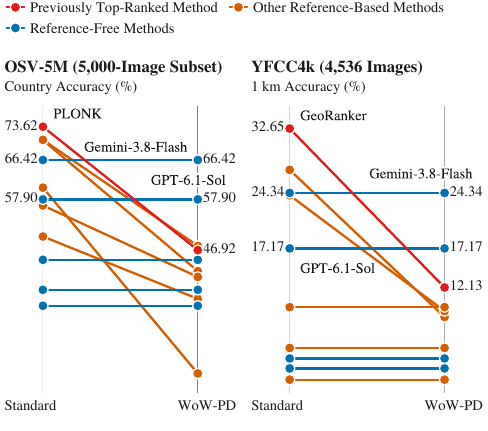}
\caption{Separating photographers flips rankings. Accuracy of 11 methods on \OSV and 10 methods on \YFCC with the standard and the photographer-disjoint (WoW-PD) reference data.}
\label{figure:ranking_flip}
\end{figure}

\section{Introduction}
\label{section:introduction}

Visual geolocation models infer where a photo was taken from its pixels alone.
This capability powers applications such as crime investigation and navigation~\cite{JLLZWDHWWY24,JPGZL25}, and it also threatens location privacy, because recent vision-language models (VLMs) can geolocate photos posted online~\cite{LDDLZSZG25,MCHDXR24,TVSV24,LZLLZXX25}.
Progress is measured on standard public benchmarks, such as the street-view benchmark OpenStreetView-5M (\OSV)~\cite{ADSABFLNRVXZL24} and the Flickr test sets \YFCC and \IMG~\cite{VJH17}.
Specialized methods train on the data paired with each benchmark (the \OSV training set or the Flickr training set \MPS~\cite{MPE18}), and retrieval-augmented methods also query a retrieval database built from these images~\cite{JLLZWDHWWY24,JPGZL25}.
We refer to the images that a method trains on or retrieves from as its reference data.
Rankings on these benchmarks therefore decide which methods count as the state of the art and which models appear to pose the largest location privacy risk.

However, these benchmarks can also reward recognizing who took a photo.
We define the photographer of an image as the account that uploaded it.
Images from one photographer often share a camera, shooting habits, and regions.
Street-view images from one vehicle likewise share camera sensors~\cite{ADSABFLNRVXZL24}.
When a test photographer also contributes reference images, a method can thus locate the test image by matching it to these images, and the benchmark cannot tell whether the method recognized the place or the photographer.
\autoref{figure:teaser} shows such a case, in which a retrieval method locates a \YFCC test image through other images of its photographer and misses it by thousands of kilometers without them.
We term this effect photographer leakage.
Some benchmarks already limit what test images share with reference images.
For instance, \OSV keeps test images at least 1 km away from training images and never splits an image sequence (a continuous series of images from one photographer) between the two sets~\cite{ADSABFLNRVXZL24}, and \MPS excludes the photographers of \IMG~\cite{MPE18}.
Nonetheless, our audit of five visual geolocation benchmarks finds that in all except \IMG, 55.95\% to 98.24\% of test images come from photographers who also appear in the paired reference data (\autoref{section:audit}).

To measure photographer leakage, we build WoW-PD (Where or Who, Photographer-Disjoint), a set of benchmark splits with photographer-disjoint reference data for \OSV and \YFCC (\autoref{section:evaluation}).
Concretely, for \OSV, we re-split the data so that each photographer contributes images to either the training set or the test set, and for \YFCC, we remove all images of its test photographers from \MPS.
For each benchmark, a size-matched random control removes the same number of random images, which separates the effect of photographers from the effect of data size.
We then retrain or re-index the specialized methods and compare them with zero-shot VLMs, which use no reference data, on the same test images.
On \OSV, separating photographers lowers the country accuracy of three methods spanning generation, retrieval, and classification by 26.62\% to 39.01\% relative to the control, whereas the control itself changes it by at most 1.24\% (\autoref{section:results}).
Similarly, on \YFCC, removing the test photographers from the retrieval database lowers the 1 km accuracy of three methods that query it by 9.04\% to 12.94\% relative to the control.
Rankings therefore flip on both benchmarks (\autoref{figure:ranking_flip}).
In particular, the generative method PLONK~\cite{DPKL25}, previously top-ranked on \OSV, falls to fourth, and the retrieval-augmented method GeoRanker~\cite{JPGZL25}, previously top-ranked on \YFCC, falls to third, while the zero-shot VLMs Gemini-3.8-Flash~\cite{gemini38flash} and GPT-6.1-Sol~\cite{gpt61sol} rise to first and second on both benchmarks.

To trace the gain that photographer leakage gives specialized methods, we take nearest-neighbor retrieval and the retrieval-augmented method G3~\cite{JLLZWDHWWY24} as examples and compare how their retrievers rank database images from the query's own photographer and from other photographers (\autoref{section:mechanism}).
On \OSV, both retrievers prefer images of the query's own photographer to other photographers' images with the same visual similarity and distance to the query location.
Specifically, with visual similarity measured by a separate DINOv2 image encoder~\cite{ODMVSKFHMEABGHHLMRSSXJMLJB24,DOMB24}, an image from another photographer needs a 0.06 to 0.08 higher similarity to be preferred equally often.
Taken together, rankings on these benchmarks partly reflect who took a photo rather than where it was taken, which misleads model selection and understates the location privacy risk of zero-shot VLMs relative to specialized methods.
We therefore recommend photographer-disjoint evaluation and release WoW-PD to support the development of geolocation methods that generalize to unseen photographers.

Our contributions are summarized as follows.
\begin{itemize}
\item We audit standard visual geolocation benchmarks and find that most of them draw the majority of test images from photographers who also contribute reference images (\autoref{section:audit}).
\item We build WoW-PD, which provides photographer-disjoint reference data for \OSV and \YFCC with size-matched random controls, and show that separating photographers lowers the accuracy of three methods on each benchmark relative to the controls and flips rankings on both benchmarks (\autoref{section:evaluation} and~\autoref{section:results}).
\item We trace the gain that photographer leakage gives the analyzed retrieval methods to images of the query's own photographer, which they prefer even at equal DINOv2 similarity and distance (\autoref{section:mechanism}).
\end{itemize}

\section{Related Work}
\label{section:related_work}

\mypara{Visual Geolocation Methods}
Visual geolocation methods differ in how they use reference data.
Classification methods divide the Earth into cells and train a classifier over them~\cite{WKP16,MPE18,HSAF24}, as does the QuadTree classifier released with the street-view benchmark OpenStreetView-5M (\OSV)~\cite{ADSABFLNRVXZL24}.
Retrieval methods return the location of the most similar image in a retrieval database~\cite{HE08,VJH17} or align images with GPS coordinates and retrieve a coordinate~\cite{CNS23}.
Generative methods instead learn a distribution over locations from the training data~\cite{DPKL25}.
Furthermore, retrieval-augmented methods first retrieve from a database and then let vision-language models (VLMs) generate or rerank candidate locations~\cite{ZZGHLMLM24,JLLZWDHWWY24,JPGZL25,LND26}, and other methods fine-tune a VLM on reasoning data built from reference data~\cite{LYJZ24,LZLTW25}.
In contrast, zero-shot VLMs such as Qwen2.5-VL-7B and Qwen2.5-VL-32B~\cite{BCLWGSDWWTZZYLWWDFXYZXCZYXL25}, Gemini-3.8-Flash~\cite{gemini38flash}, and GPT-6.1-Sol~\cite{gpt61sol} use no reference data, so photographer leakage can raise only the accuracy of the methods above.

\mypara{Benchmarks and Their Splits}
Common benchmarks include the Flickr test sets Im2GPS~\cite{HE08}, \IMG and \YFCC~\cite{VJH17}, and YFCC26k~\cite{MPE18}, the street-view benchmark \OSV, and the test set MP16-Reason-Test of the fine-tuned geolocation VLM GLOBE~\cite{LZLTW25}.
These Flickr test sets are usually paired with the training set \MPS~\cite{LSGIJ17} or its subset \MPP~\cite{JLLZWDHWWY24}.
In addition, the split of \OSV separates test from training images by distance and image sequence, and MP16-Reason-Test covers locations absent from its training set.
Some works separate photographers.
\IMG and \YFCC were built without photographers shared with their original training data~\cite{VJH17}.
\MPS excludes the photographers of the Im2GPS and \IMG test sets~\cite{MPE18}, and the transformer-based method TransLocator excludes the images of its validation and test photographers from \MPS during training~\cite{PNGCC22}.
Although the classification method PIGEON quantifies the location overlap between test images and training data~\cite{HSAF24}, none of these works quantifies how many test images share a photographer with the reference data.

\mypara{Leakage, Shortcuts, and Group Splits}
Locations, near-duplicate images, or evaluation information shared between training and test data inflate scores and can change conclusions.
For instance, online mapping datasets overlap geographically between training and evaluation samples, and geographically disjoint splits lower the performance of several methods and change design conclusions~\cite{LFSH24}.
Similarly, supervised pretraining and oracle model selection leak test-domain information in domain generalization, where a corrected protocol changes the leaderboard~\cite{YZXLHC24}.
The CIFAR test sets also contain near-duplicates of training images~\cite{BD20}.
Success that relies on such shared factors is a form of shortcut learning~\cite{GJMZBBW20}.
Other fields therefore split their data by identity or group.
Person re-identification~\cite{ZSTWWT15}, face recognition~\cite{CSXPZ18}, and speaker verification~\cite{CNZ18} keep training and test identities disjoint.
Likewise, medical imaging studies split data by subject to avoid inflated accuracy~\cite{YAHMSGTCD21}, and spatial data call for block cross-validation~\cite{RBCBEGHLSTWWHD17}.
Yet, most of the common visual geolocation benchmarks do not separate test images from reference data by photographer (\autoref{section:audit}), so we treat the photographer as the unit to group by.

\mypara{Geolocation and Location Privacy}
VLMs can infer locations from photos posted online, a capability studied as a location privacy risk~\cite{LDDLZSZG25,MCHDXR24,TVSV24,LZLLZXX25,ZWZLSBZ25}.
Additionally, defenses treat VLMs as the main threat~\cite{LJXQC26}.
Judging which kind of model poses the larger location privacy risk relies on benchmark rankings, and for photos whose photographer is absent from the reference data, photographer leakage in these rankings understates the risk of zero-shot VLMs relative to specialized methods.

\section{Photographer Overlap in Benchmarks}
\label{section:audit}

This section examines how widespread photographer overlap is in current benchmarks.

\mypara{Audit Protocol}
We obtain each image's photographer from public metadata (the Flickr account for the Flickr test sets and the Mapillary account for \OSV).
For each benchmark, we measure photographer overlap, the share of test images whose photographer has images in the paired reference data (\autoref{table:benchmark_audit}).
Moreover, we report the number of test photographers and the share of test images from the 10 photographers with the most test images (Top-10 Share), because accuracy is uncertain when few photographers contribute most test images.
To gauge how widely each benchmark is used, we also count how many of the 92 geolocation method papers collected from arXiv (January 2024 to October 2026) report numerical results on it (literature census in~\autoref{appendix:literature_census}).

\begin{table*}[t]
\centering
\caption{Photographer overlap of the audited visual geolocation benchmarks. Papers counts how many of the 92 collected papers report results on each benchmark, alone or with other tracked benchmarks (literature census in~\autoref{appendix:literature_census}).}
\label{table:benchmark_audit}
\scalebox{0.90}{
\begin{tabular}{@{}lrrrlrrr@{}}
\toprule
 & & & & & & \multicolumn{2}{c}{Papers} \\
\cmidrule(l){7-8}
Benchmark & Test Images & Photographers & Top-10 Share & Reference Data & Overlap & Only This & With Others \\
\midrule
\OSV & 210,122 & 5,318 & 18.27\% & \OSV Training Set & 98.24\% & 2 & 11 \\
\YFCC & 4,536 & 3,481 & 2.98\% & \MPS & 75.29\% & 0 & 23 \\
YFCC26k & 25,600 & 13,018 & 2.46\% & \MPS & 70.21\% & 0 & 5 \\
MP16-Reason-Test & 12,000 & 9,786 & 0.48\% & MP16-Reason & 55.95\% & 0 & 3 \\
\IMG & 2,997 & 198 & 32.63\% & \MPS & 0.00\% & 6 & 36 \\
\bottomrule
\end{tabular}
}
\end{table*}

\mypara{Photographer Overlap}
In four of the five audited benchmarks, most test images come from photographers who also contribute reference images (\autoref{table:benchmark_audit}).
Specifically, photographer overlap is 98.24\% on \OSV, 75.29\% on \YFCC, 70.21\% on YFCC26k, and 55.95\% on MP16-Reason-Test.
\MPS even contains 534 of the \YFCC test images themselves (11.77\%).
The overlap on \OSV remains high because neither the 1 km buffer nor the rule against splitting image sequences separates photographers.
Indeed, the \OSV authors note that they could not separate images uploaded by the same photographer on different days~\cite{ADSABFLNRVXZL24}.
Even among the 2.59\% of \OSV test images that lie at least 100 km from every training image, photographer overlap is 86.90\%.

\mypara{Im2GPS3k and Benchmark Use}
\IMG is the only audited benchmark without photographer overlap, because \MPS excludes the photographers of \IMG~\cite{MPE18}.
However, its 2,997 test images were taken by only 198 photographers at 1,941 distinct coordinates, whereas the 4,536 test images of \YFCC were taken by 3,481 photographers at 4,426 distinct coordinates.
The accuracy of a method on \IMG thus depends heavily on these few photographers and locations.
\IMG alone therefore cannot replace the benchmarks with photographer overlap.
These benchmarks also remain widely used.
Among the 92 papers, 23 report results on \YFCC and 13 on \OSV, and 29 of the 42 papers that report results on \IMG also do so on at least one benchmark with photographer overlap.
We therefore build photographer-disjoint reference data for \OSV and \YFCC, the two most used of these benchmarks (\autoref{section:evaluation}).

\begin{tcolorbox}[colback=gray!10!white, size=title,breakable,boxsep=1mm,colframe=white,before={\vskip1mm}, after={\vskip0mm}]
\emph{\textbf{Takeaways:}
Most of the audited benchmarks draw the majority of their test images from photographers who also contribute reference images, and the only exception, \IMG, has too few photographers and locations to replace them.}
\end{tcolorbox}

\section{WoW-PD: Photographer-Disjoint Splits}
\label{section:evaluation}

The WoW-PD splits provide photographer-disjoint reference data for \OSV, \YFCC, and YFCC26k, each with a size-matched random control.
They consist of image ID lists.
We evaluate on \OSV and \YFCC and release the YFCC26k lists for future use.
All reference data conditions share the same test images, so accuracy differences between them come only from the reference data.
Furthermore, WoW-PD includes our \OSV test split to evaluate future methods on photographers absent from their reference data.

\subsection{Re-Splitting OSV-5M by Photographer}
\label{subsection:osv_resplit}

We re-split \OSV so that each photographer, and hence each image sequence, contributes images to only one of the training, validation, and test splits.
We keep the official 1 km buffer and match the country distribution of the official test set, so that apart from separating photographers, our split stays close to the official one.
Specifically, we greedily select test photographers, with the number of official test images in each country as an upper limit, and allocate each country's test images across the cells of the official \OSV quadtree.
We then remove training images within 1 km of any test or validation image (\autoref{algorithm:split}).
As a result, the photographer-disjoint training set has 4,020,558 images, which is 82.14\% of the official training set.
The test split has 209,386 images from 2,960 photographers, and the validation split has 3,924 images.
The total variation distance between the country distributions of our test split and the official test set is 0.35\%.
For comparison, the control training set draws as many images as the photographer-disjoint training set uniformly at random from the official training set.
Thus, the control differs from the official training set only in size.
Since all three training sets contain only official training images, we evaluate them on 126,972 of the official test images in our test split (re-split details in~\autoref{appendix:split_details}).

\begin{algorithm}[t]
\caption{Photographer-disjoint re-split of \OSV.}
\label{algorithm:split}
\small
\begin{algorithmic}[1]
\Require pool of official training and test images with photographer, country, and cell labels; quota $Q_k$ of each country $k$, equal to its number of official test images; (country, cell) pairs $p$ with $t_p > 0$ official test images; stopping threshold $\lambda$
\Ensure photographer-disjoint splits $\mathcal{I}_{\mathrm{train}}$, $\mathcal{I}_{\mathrm{val}}$, and $\mathcal{I}_{\mathrm{test}}$
\State $S \gets \emptyset$ \Comment{test photographers}
\While{some quota is unfilled}
  \State $u^\ast \gets \arg\max_{u \notin S} s_u$ with $s_u = g_u / (n_u - g_u + 1)$, where $g_u$ is the remaining quota that photographer $u$ fills and $n_u$ its number of images
  \Statex \Comment{quota filled per image left out of all splits}
  \State \textbf{if} $s_{u^\ast} < \lambda$ \textbf{then break} \Comment{keeps the training set large}
  \State $S \gets S \cup \{u^\ast\}$
\EndWhile
\State $x_p \gets \min(a_p, \theta_k t_p)$ for each pair $p$ of country $k$, where $a_p$ counts the images of $S$ in $p$ and $\theta_k$ gives $\sum_p x_p = \min(Q_k, \sum_p a_p)$ \Comment{follows official cell counts}
\For{$u \in S$ in reverse order of selection}
  \State \textbf{if} every $x_p$ stays filled without $u$ \textbf{then} $S \gets S \setminus \{u\}$
  \Statex \Comment{returns unneeded photographers}
\EndFor
\State $\mathcal{I}_{\mathrm{test}} \gets$ $x_p$ images of $S$ in each pair $p$, official test images first
\State $\mathcal{I}_{\mathrm{val}} \gets$ output of lines 1 to 11 on the official training images of the remaining photographers, with 2\% of the quotas
\State $\mathcal{I}_{\mathrm{train}} \gets$ official training images of the photographers in neither split that lie more than 1 km from $\mathcal{I}_{\mathrm{test}} \cup \mathcal{I}_{\mathrm{val}}$
\end{algorithmic}
\end{algorithm}

\subsection{Removing Photographers from MP-16}
\label{subsection:mp16_removal}

On \YFCC, the retrieval-augmented methods G3~\cite{JLLZWDHWWY24} and GeoRanker~\cite{JPGZL25} retrieve from \MPP, a 4,122,119-image subset of \MPS, which serves as the original database.
This database contains images from 2,437 of the 3,481 \YFCC test photographers.
We therefore remove all 283,095 of their images, which yields the photographer-removed database.
The control database first removes the 467 \YFCC test images that \MPP already contains and then removes random images until it has the same size as the photographer-removed database (3,839,024 images).
Moreover, WoW-PD applies the same removal to the full \MPS list and provides the corresponding lists for YFCC26k.
We swap only the retrieval database on \YFCC and keep all model weights fixed, which isolates the leakage that enters through retrieval.

\subsection{Evaluation Protocol}
\label{subsection:protocol}

\mypara{Methods on OSV-5M}
On \OSV, we evaluate 11 methods in three groups, according to how they use reference data.
We retrain the generative method PLONK~\cite{DPKL25} and the QuadTree classifier, a classification baseline of \OSV~\cite{ADSABFLNRVXZL24}, on each training set, keeping the classifier's backbone frozen.
Nearest-neighbor retrieval, which copies the location of the most similar database image,\footnote{Nearest-neighbor retrieval uses StreetCLIP~\cite{HAS23} features on \OSV and the G3 encoder on \YFCC.} and the retrieval-augmented methods G3~\cite{JLLZWDHWWY24}, GeoRanker~\cite{JPGZL25}, and GeoSearch~\cite{LND26} use each training set as their retrieval database.
The fine-tuned geolocation VLM GLOBE~\cite{LZLTW25} and the zero-shot VLMs Qwen2.5-VL-7B, Qwen2.5-VL-32B~\cite{BCLWGSDWWTZZYLWWDFXYZXCZYXL25}, Gemini-3.8-Flash~\cite{gemini38flash}, and GPT-6.1-Sol~\cite{gpt61sol} keep fixed weights and use none of the training sets.
We compare PLONK, the QuadTree classifier, and nearest-neighbor retrieval on 126,972 of the official test images in our test split.
Because the closed-source VLMs, G3, and GeoRanker are costly to run, we compare all 11 methods on a 5,000-image subset of these images.
We draw this subset by sorting the images by country and photographer and taking 5,000 equally spaced positions from a random start, so that every image has the same chance of selection.

\mypara{Methods on YFCC4k}
On \YFCC, we evaluate 10 methods on all 4,536 test images.
G3, GeoRanker, and nearest-neighbor retrieval retrieve from \MPP, so we evaluate them with the original, control, and photographer-removed databases (\autoref{subsection:mp16_removal}).
The other seven methods keep fixed weights and use none of these databases.
They are the four zero-shot VLMs and three methods whose weights are trained on \MPS data (GLOBE, the encoder of GeoSearch, and the image-to-GPS retrieval method GeoCLIP~\cite{CNS23}).
GeoSearch retrieves from the \OSV training set on every benchmark.
Since this database contains no Flickr images, there are no \YFCC test photographers to remove from it.
Thus, we swap its retrieval database only on \OSV, and on \YFCC we evaluate only its encoder.
In~\autoref{figure:ranking_flip} and~\autoref{figure:prediction_examples}, a method is reference-based on a benchmark if it trains on or retrieves from the reference data of that benchmark, and reference-free otherwise.
In figures and tables, Standard denotes the official training set or the original database, and WoW-PD the photographer-disjoint training set or the photographer-removed database.

\mypara{Metrics and Confidence Intervals}
We compare methods by country accuracy on \OSV and by 1 km accuracy on \YFCC.
In addition,~\autoref{table:main_results} lists for \OSV a score that decreases from 5,000 toward 0 as the distance error grows (GeoScore), the average distance error, and the region and city accuracy~\cite{ADSABFLNRVXZL24}.
It also reports the accuracy within 25, 200, 750, and 2,500 km on \YFCC~\cite{VJH17}.
We measure the effect of separating photographers on a method as its accuracy with the control minus its accuracy with the photographer-disjoint reference data.
Because test images of one photographer are not independent, we compute 95\% confidence intervals with a bootstrap over test photographers rather than images.
Concretely, we draw test photographers with replacement 10,000 times, keep all test images of each drawn photographer, recompute the results on each resample, and take the 2.5th and 97.5th percentiles as the interval.
All methods and conditions share the same resamples, so each difference between them is computed on the same photographers.
Prompts, model versions, and the remaining metrics are given in~\autoref{appendix:implementation}, and complete results in~\autoref{appendix:complete_results}.

\section{Separating Photographers Flips Rankings}
\label{section:results}

\begin{table*}[t]
\centering
\caption{Results under each training set or retrieval database on 126,972 \OSV test images and all 4,536 \YFCC test images.}
\label{table:main_results}
\renewcommand{\arraystretch}{1.08}
\scalebox{0.73}{
\begin{tabular}{@{}lrrrrr@{\hskip 2.5em}lrrrrr@{}}
\toprule
\multicolumn{6}{c}{\OSV} & \multicolumn{6}{c}{\YFCC} \\
\cmidrule(r){1-6} \cmidrule(l){7-12}
 & & & \multicolumn{3}{c}{Accuracy (\%)} & & \multicolumn{5}{c}{Accuracy (\%)} \\
\cmidrule(lr){4-6} \cmidrule(l){8-12}
Training Set & GeoScore & Distance (km) & Country & Region & City & Retrieval Database & 1 km & 25 km & 200 km & 750 km & 2,500 km \\
\midrule
\multicolumn{6}{@{}l}{\emph{PLONK}} & \multicolumn{6}{l@{}}{\emph{GeoRanker}} \\
Official & 3,581.99 & 1,263.58 & 72.72 & 40.25 & 4.81 & Original & 32.65 & 43.41 & 53.97 & 69.82 & 82.32 \\
Control & 3,599.01 & 1,274.43 & 73.17 & 41.44 & 5.12 & Control & 25.07 & 37.17 & 48.77 & 66.42 & 80.22 \\
\rowcolor{gray!20} Photographer-Disjoint & 2,226.51 & 2,913.34 & 46.55 & 10.14 & 0.19 & Photographer-Removed & 12.13 & 24.36 & 38.21 & 59.57 & 77.29 \\
\textcolor[HTML]{E31A1C}{Change from Control} & \textcolor[HTML]{E31A1C}{$-$1,372.50} & \textcolor[HTML]{E31A1C}{$+$1,638.91} & \textcolor[HTML]{E31A1C}{$-$26.62} & \textcolor[HTML]{E31A1C}{$-$31.30} & \textcolor[HTML]{E31A1C}{$-$4.93} & \textcolor[HTML]{E31A1C}{Change from Control} & \textcolor[HTML]{E31A1C}{$-$12.94} & \textcolor[HTML]{E31A1C}{$-$12.81} & \textcolor[HTML]{E31A1C}{$-$10.56} & \textcolor[HTML]{E31A1C}{$-$6.86} & \textcolor[HTML]{E31A1C}{$-$2.93} \\
\midrule
\multicolumn{6}{@{}l}{\emph{Nearest-Neighbor Retrieval}} & \multicolumn{6}{l@{}}{\emph{Nearest-Neighbor Retrieval}} \\
Official & 3,403.81 & 1,612.34 & 70.06 & 42.72 & 8.34 & Original & 27.31 & 35.52 & 43.92 & 59.44 & 73.92 \\
Control & 3,378.03 & 1,645.39 & 69.58 & 42.02 & 8.12 & Control & 18.87 & 28.33 & 37.59 & 54.78 & 71.05 \\
\rowcolor{gray!20} Photographer-Disjoint & 2,061.84 & 3,188.79 & 42.38 & 9.01 & 0.27 & Photographer-Removed & 8.31 & 17.00 & 27.38 & 46.96 & 66.78 \\
\textcolor[HTML]{E31A1C}{Change from Control} & \textcolor[HTML]{E31A1C}{$-$1,316.19} & \textcolor[HTML]{E31A1C}{$+$1,543.40} & \textcolor[HTML]{E31A1C}{$-$27.20} & \textcolor[HTML]{E31A1C}{$-$33.01} & \textcolor[HTML]{E31A1C}{$-$7.85} & \textcolor[HTML]{E31A1C}{Change from Control} & \textcolor[HTML]{E31A1C}{$-$10.56} & \textcolor[HTML]{E31A1C}{$-$11.33} & \textcolor[HTML]{E31A1C}{$-$10.21} & \textcolor[HTML]{E31A1C}{$-$7.83} & \textcolor[HTML]{E31A1C}{$-$4.28} \\
\midrule
\multicolumn{6}{@{}l}{\emph{QuadTree Classifier}} & \multicolumn{6}{l@{}}{\emph{G3}} \\
Official & 3,100.43 & 2,342.56 & 60.50 & 34.75 & 3.51 & Original & 24.03 & 33.91 & 44.07 & 61.09 & 75.60 \\
Control & 3,030.67 & 2,438.78 & 59.25 & 32.78 & 3.27 & Control & 18.12 & 28.95 & 39.57 & 57.78 & 73.57 \\
\rowcolor{gray!20} Photographer-Disjoint & 1,136.42 & 5,847.86 & 20.25 & 3.26 & 0.07 & Photographer-Removed & 9.08 & 18.85 & 30.75 & 51.15 & 70.19 \\
\textcolor[HTML]{E31A1C}{Change from Control} & \textcolor[HTML]{E31A1C}{$-$1,894.25} & \textcolor[HTML]{E31A1C}{$+$3,409.08} & \textcolor[HTML]{E31A1C}{$-$39.01} & \textcolor[HTML]{E31A1C}{$-$29.52} & \textcolor[HTML]{E31A1C}{$-$3.20} & \textcolor[HTML]{E31A1C}{Change from Control} & \textcolor[HTML]{E31A1C}{$-$9.04} & \textcolor[HTML]{E31A1C}{$-$10.10} & \textcolor[HTML]{E31A1C}{$-$8.82} & \textcolor[HTML]{E31A1C}{$-$6.64} & \textcolor[HTML]{E31A1C}{$-$3.37} \\
\bottomrule
\end{tabular}
}
\end{table*}

\begin{table*}[t]
\centering
\caption{Accuracy and rank of every method under each reference data condition, on the 5,000-image subset of \OSV and on \YFCC.}
\label{table:rankings}
\renewcommand{\arraystretch}{1.08}
\scalebox{0.74}{
\begin{tabular}{@{}lrrr>{\columncolor{gray!20}}r>{\columncolor{gray!20}}r@{\hskip 1.5em}lrrr>{\columncolor{gray!20}}r>{\columncolor{gray!20}}r@{}}
\toprule
\multicolumn{6}{c}{\OSV, Country Accuracy (\%)} & \multicolumn{6}{c}{\YFCC, 1 km Accuracy (\%)} \\
\cmidrule(r){1-6} \cmidrule(l){7-12}
Method & Standard & Rank & Control & WoW-PD & Rank & Method & Standard & Rank & Control & WoW-PD & Rank \\
\midrule
PLONK & 73.62 & 1 & 73.74 & 46.92 & \textcolor[HTML]{E31A1C}{4$\downarrow$} & GeoRanker & 32.65 & 1 & 25.07 & 12.13 & \textcolor[HTML]{E31A1C}{3$\downarrow$} \\
GeoRanker & 70.78 & 2 & 70.14 & 47.82 & \textcolor[HTML]{E31A1C}{3$\downarrow$} & Nearest-Neighbor Retrieval & 27.31 & 2 & 18.87 & 8.31 & \textcolor[HTML]{E31A1C}{6$\downarrow$} \\
Nearest-Neighbor Retrieval & 70.74 & 3 & 70.44 & 42.38 & \textcolor[HTML]{E31A1C}{6$\downarrow$} & Gemini-3.8-Flash & 24.34 & 3 & -- & 24.34 & \textcolor[HTML]{009E73}{1$\uparrow$} \\
Gemini-3.8-Flash & 66.42 & 4 & -- & 66.42 & \textcolor[HTML]{009E73}{1$\uparrow$} & G3 & 24.03 & 4 & 18.12 & 9.08 & \textcolor[HTML]{E31A1C}{5$\downarrow$} \\
QuadTree Classifier & 60.44 & 5 & 58.84 & 20.22 & \textcolor[HTML]{E31A1C}{11$\downarrow$} & GPT-6.1-Sol & 17.17 & 5 & -- & 17.17 & \textcolor[HTML]{009E73}{2$\uparrow$} \\
GPT-6.1-Sol & 57.90 & 6 & -- & 57.90 & \textcolor[HTML]{009E73}{2$\uparrow$} & GeoCLIP & 9.59 & 6 & -- & 9.59 & \textcolor[HTML]{009E73}{4$\uparrow$} \\
G3 & 56.58 & 7 & 55.44 & 41.12 & \textcolor[HTML]{666666}{7\phantom{$\uparrow$}} & GLOBE & 4.32 & 7 & -- & 4.32 & \textcolor[HTML]{666666}{7\phantom{$\uparrow$}} \\
GeoSearch & 49.86 & 8 & 49.18 & 36.38 & \textcolor[HTML]{E31A1C}{9$\downarrow$} & Qwen2.5-VL-32B & 2.95 & 8 & -- & 2.95 & \textcolor[HTML]{666666}{8\phantom{$\uparrow$}} \\
GLOBE & 44.82 & 9 & -- & 44.82 & \textcolor[HTML]{009E73}{5$\uparrow$} & Qwen2.5-VL-7B & 1.68 & 9 & -- & 1.68 & \textcolor[HTML]{666666}{9\phantom{$\uparrow$}} \\
Qwen2.5-VL-7B & 38.32 & 10 & -- & 38.32 & \textcolor[HTML]{009E73}{8$\uparrow$} & GeoSearch Encoder & 0.22 & 10 & -- & 0.22 & \textcolor[HTML]{666666}{10\phantom{$\uparrow$}} \\
Qwen2.5-VL-32B & 34.88 & 11 & -- & 34.88 & \textcolor[HTML]{009E73}{10$\uparrow$} & \multicolumn{6}{c}{} \\
\bottomrule
\end{tabular}
}
\end{table*}

\subsection{Separating Photographers Lowers Accuracy}
\label{subsection:accuracy_drop}

\mypara{Results on OSV-5M}
Replacing the control training set with the photographer-disjoint training set lowers the country accuracy of three methods spanning generation, retrieval, and classification by 26.62\% to 39.01\% on 126,972 of the official \OSV test images in our test split (\autoref{table:main_results}).
Specifically, the country accuracy of PLONK falls from 73.17\% to 46.55\% ($-$26.62\%, 95\% confidence interval [$-$30.41\%, $-$22.84\%]).
Nearest-neighbor retrieval falls from 69.58\% to 42.38\% ($-$27.20\% [$-$30.45\%, $-$23.66\%]), and the QuadTree classifier falls from 59.25\% to 20.25\% ($-$39.01\% [$-$42.43\%, $-$35.60\%]).
In contrast, the control training set changes country accuracy by at most 1.24\% relative to the official training set, although it contains 17.86\% fewer images.
Hence, the drop comes from separating photographers rather than from the smaller training set.
All six methods whose training set or retrieval database we swap show the drop on the 5,000-image subset, ranging from 12.80\% for GeoSearch to 38.62\% for the QuadTree classifier (\autoref{table:rankings}).

Separating photographers also leaves some cells without training images, so part of the drop could come from reduced geographic coverage.
Nonetheless, 95,303 of the 126,972 test images lie in cells that still contain training images.
On these images, the drop remains 24.84\% for PLONK, 25.35\% for nearest-neighbor retrieval, and 36.20\% for the QuadTree classifier.
The drop therefore mainly comes from separating photographers rather than from reduced coverage.
Conversely, the photographer-disjoint training set changes country accuracy by only $+$0.27\% to $+$2.04\% relative to the control on 77,148 of the official test images whose photographers remain in this training set.

\mypara{Results on YFCC4k}
Likewise, on \YFCC, replacing the control database with the photographer-removed database lowers the 1 km accuracy of the three methods that query \MPP by 9.04\% to 12.94\%.
GeoRanker falls from 25.07\% to 12.13\% ($-$12.94\%), nearest-neighbor retrieval from 18.87\% to 8.31\% ($-$10.56\%), and G3 from 18.12\% to 9.08\% ($-$9.04\%).
By contrast, on \IMG, whose photographers are absent from \MPS, removing the images of the \YFCC test photographers from the original database barely changes the 1 km accuracy of G3, from 13.78\% to 13.85\%.
The drop is thus specific to test images whose photographers are removed from the database.
Moreover, these drops on \YFCC are lower bounds on the effect of photographer leakage, because the encoder that these methods use for retrieval and the reranker of GeoRanker are still trained on \MPP.

\subsection{Ranking Flips on Benchmarks}
\label{subsection:ranking_flips}

\begin{figure}[t]
\centering
\includegraphics[width=\columnwidth]{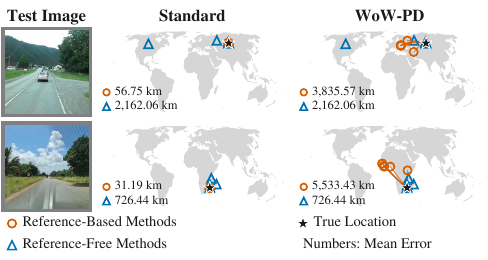}
\caption{Predictions of all 11 methods for two \OSV test images with the standard and the WoW-PD reference data.}
\label{figure:prediction_examples}
\end{figure}

\mypara{Rankings on OSV-5M}
We next rank all 11 methods on the 5,000-image subset of \OSV by country accuracy, once with the official and once with the photographer-disjoint reference data (\autoref{table:rankings} and~\autoref{figure:ranking_flip}, left).
With the official reference data, the specialized methods PLONK (73.62\%), GeoRanker (70.78\%), and nearest-neighbor retrieval (70.74\%) take the top three places, while the zero-shot VLMs Gemini-3.8-Flash (66.42\%) and GPT-6.1-Sol (57.90\%) rank fourth and sixth.
However, with photographers separated, Gemini-3.8-Flash and GPT-6.1-Sol rise to first and second, PLONK falls to fourth (46.92\%), and the QuadTree classifier from fifth to eleventh.
The two closed-source VLMs also outrank all other methods in GeoScore and in the accuracy within 1, 25, 200, 750, and 2,500 km (complete results in~\autoref{appendix:complete_results}).
The country accuracy of PLONK minus that of Gemini-3.8-Flash changes from $+$7.20\% [3.82\%, 10.66\%] to $-$19.50\% [$-$22.92\%, $-$16.34\%], a reversal beyond sampling error.
For example, in~\autoref{figure:prediction_examples}, the methods that use reference data locate two \OSV test images with the official reference data but miss them by thousands of kilometers with the photographer-disjoint reference data, whereas the predictions of the other methods stay the same.

\mypara{Rankings on YFCC4k}
Similarly, rankings by 1 km accuracy flip on \YFCC when the original database is replaced with the photographer-removed database (\autoref{figure:ranking_flip}, right).
GeoRanker falls from first to third, nearest-neighbor retrieval from second to sixth, and G3 from fourth to fifth, whereas Gemini-3.8-Flash rises from third to first and GPT-6.1-Sol from fifth to second.
At 25 and 200 km, Gemini-3.8-Flash and GPT-6.1-Sol likewise end up first and second.
The seven methods that use none of these databases keep their accuracy and appear as flat lines.

\begin{figure*}[t]
\centering
\includegraphics[width=\textwidth]{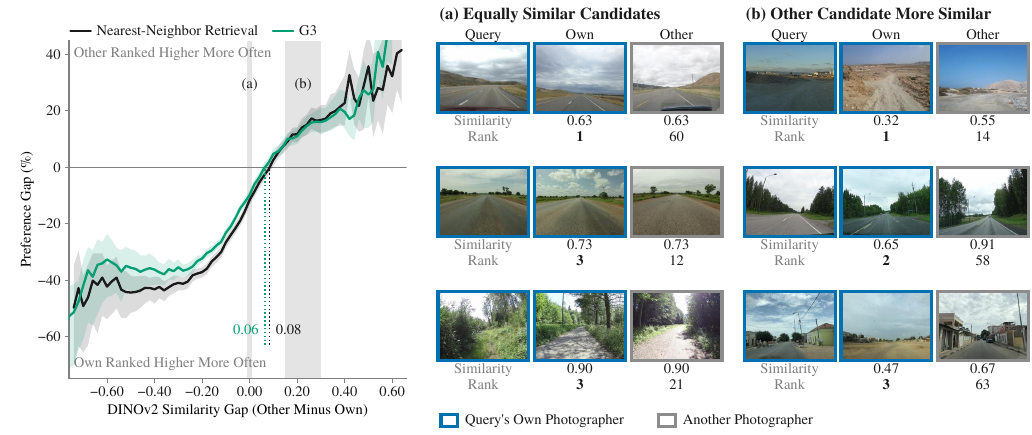}
\caption{Preference gap versus similarity gap on \OSV with 95\% confidence bands (left) and example pairs from ranges (a) and (b) with their DINOv2 similarity to the query and rank by nearest-neighbor retrieval (right).}
\label{figure:similarity_gap}
\end{figure*}

The flip appears even without any change to the retrieval database.
We split the \YFCC test images by whether their photographer has images in the original database and leave out the 467 test images that the database itself contains.
We compare the 1 km accuracy of GeoRanker and Gemini-3.8-Flash within each group.
GeoRanker leads Gemini-3.8-Flash by 6.35\% on the 2,644 test images whose photographers are in the database, whereas it trails by 9.47\% on the 1,425 test images whose photographers are absent.
The two gaps thus differ by 15.83\% [12.74\%, 18.98\%].

Since the zero-shot VLMs stay flat in both panels of~\autoref{figure:ranking_flip}, the ranking flips come entirely from the specialized methods losing the gain that photographer leakage gives them.

\begin{tcolorbox}[colback=gray!10!white, size=title,breakable,boxsep=1mm,colframe=white,before={\vskip1mm}, after={\vskip0mm}]
\emph{\textbf{Takeaways:}
Separating photographers lowers the accuracy of three specialized methods on each benchmark relative to the size-matched random controls and flips the rankings on both benchmarks, while two zero-shot VLMs rise to first and second.}
\end{tcolorbox}

\section{Tracing the Leakage Gain}
\label{section:mechanism}

To trace the gain that photographer leakage gives specialized methods, we take nearest-neighbor retrieval and the retrieval-augmented method G3 as examples and examine how their retrievers rank database images.
We first identify the images that carry the gain (\autoref{subsection:own_photographer}) and then test whether the retrievers prefer these images even at equal visual similarity (\autoref{subsection:beyond_similarity}).
Implementation details of both analyses are given in~\autoref{appendix:implementation}.

\subsection{Matching the Query's Photographer}
\label{subsection:own_photographer}

On \OSV, the top-ranked neighbor of nearest-neighbor retrieval comes from the query's own photographer for 57.90\% of the 124,822 queries whose photographer is in the database.
If the retriever picked among the database images at about the same distance from the query as its top-ranked neighbor regardless of who took them, this share would be only 32.33\%.
Moreover, the preference extends to distant neighbors, since 45.08\% of the top-ranked neighbors that lie 200 to 1,000 km from the query come from the query's own photographer against an expected 10.09\%.
Similarly, on \YFCC, the observed share is 23.22\% against an expected 14.22\%, computed over the 2,644 queries whose photographer, but not the image itself, is in the database.

To measure how much of the gain these neighbors carry, we replace every top-ranked neighbor from the query's own photographer with the highest-ranked neighbor from another photographer in the same retrieved list, which amounts to removing all images of this photographer from the database.
The replacement changes the top-ranked neighbor of 62,915 \OSV queries, whose country accuracy drops from 93.07\% to 48.34\%.
For 70.98\% of these queries, other photographers have database images within 50 km, yet the new top-ranked neighbor lies that close for only 2.88\%.
Country accuracy falls from 68.34\% to 43.96\% on the 115,461 queries whose retrieved list contains a neighbor from another photographer, close to the 39.65\% obtained with the photographer-disjoint training set as the database.
Likewise, 568 of the 2,644 \YFCC queries have their top-ranked neighbor replaced, which lowers their 1 km accuracy from 78.17\% to 5.28\%.
The 1 km accuracy of the 2,598 queries with a neighbor from another photographer falls from 24.10\% to 8.16\%, which nearly equals the 8.20\% with the photographer-removed database.
Hence, removing only the images of the query's own photographer accounts for most of the drop from separating photographers.

\subsection{Preference Beyond Visual Similarity}
\label{subsection:beyond_similarity}

Images from one photographer often look alike, so we test whether the preference for the query's own photographer remains when visual similarity is equal.
Specifically, we measure visual similarity with a fixed DINOv2 ViT-L/14 encoder with registers~\cite{ODMVSKFHMEABGHHLMRSSXJMLJB24,DOMB24}, which is independent of the encoders of both retrievers, and refer to the resulting similarity as DINOv2 similarity.
Among each retriever's top 100 candidates for each \OSV query, we pair every image from the query's own photographer with every image from another photographer when the two have a matched distance to the query location and lie both inside or both outside the query's country.
The similarity gap of a pair is the DINOv2 similarity between the query and the other photographer's image minus that for the image of the query's own photographer.
Analogously, the preference gap is the share of pairs in which the retriever ranks the other photographer's image higher minus the share in which it ranks the image of the query's own photographer higher, averaged over queries (\autoref{figure:similarity_gap}).

Both retrievers prefer images of the query's own photographer even at a similarity gap of zero, where the preference gap is $-$11.57\% for nearest-neighbor retrieval and $-$9.80\% for G3.
An image from another photographer needs a DINOv2 similarity that is 0.08 higher for nearest-neighbor retrieval and 0.06 higher for G3 to be preferred equally often.
For instance,~\autoref{figure:similarity_gap} shows queries for which nearest-neighbor retrieval ranks an image of the query's own photographer among its top three, ahead of another photographer's image that is equally similar to the query (a) or even 0.15 to 0.30 more similar (b).
Nearest-neighbor retrieval still ranks the image of the query's own photographer higher in 43.73\% of the pairs in range (b).
Moreover, the G3 encoder is trained only on Flickr images from \MPP, yet it shows this preference for \OSV photographers it never saw.
Images of one photographer share features beyond DINOv2 similarity and distance, such as the camera and shooting habits.
The retrievers thus take a shortcut~\cite{GJMZBBW20}, matching who took a photo rather than where it was taken.

\begin{tcolorbox}[colback=gray!10!white, size=title,breakable,boxsep=1mm,colframe=white,before={\vskip1mm}, after={\vskip0mm}]
\emph{\textbf{Takeaways:}
The gain that photographer leakage gives the analyzed retrieval methods comes mainly from images of the query's own photographer, which they prefer even at equal DINOv2 similarity and distance.}
\end{tcolorbox}

\section{Conclusion}
\label{section:conclusion}

Most standard visual geolocation benchmarks draw 55.95\% to 98.24\% of their test images from photographers who also appear in the reference data.
With the WoW-PD splits and size-matched random controls, separating photographers lowers the country accuracy of three methods on \OSV by 26.62\% to 39.01\% relative to the control, whereas the control itself changes it by at most 1.24\%.
Rankings consequently flip on both \OSV and \YFCC, as the previously top-ranked methods (PLONK and GeoRanker) fall to fourth and third, respectively, while two zero-shot VLMs rise to first and second.
Moreover, the gain that photographer leakage gives the analyzed retrieval methods comes mainly from images of the query's own photographer, which they prefer even at equal DINOv2 similarity and distance.
Hence, standard benchmark rankings partly reflect who took a photo rather than where it was taken, which misleads model selection and understates the location privacy risk of zero-shot VLMs relative to specialized methods.
We therefore recommend photographer-disjoint evaluation and release the WoW-PD splits to support the development of geolocation methods that generalize to unseen photographers.

\bibliographystyle{plain}
\bibliography{normal_generated_py3}

\appendix

\section{Literature Census}
\label{appendix:literature_census}

\mypara{Paper Collection}
We query the arXiv API for papers first submitted between January 1, 2024 and October 31, 2026 whose title or abstract contains geolocation or one of four spelling variants (geolocalization, geo-localization, geo-location, and geolocalisation) in six categories (cs.CV, cs.LG, cs.AI, cs.CL, cs.MM, and cs.IR).
These queries return 374 papers.
Of these, we include papers that propose or evaluate methods that predict where a single ground-level image was taken.
Keyword rules on the title and abstract first exclude 180 papers, most of which concern cross-view, aerial, or satellite imagery.
One of the remaining 194 papers was withdrawn.
Two authors then independently read the other 193 papers and, resolving disagreements by discussion, include 92 of them in the census.

\mypara{Benchmark Use}
A paper uses a benchmark if it reports numerical results on the official test set of the benchmark or on a subset of it.
Results on re-collected, re-split, or relabeled versions of a benchmark do not count as uses of that benchmark, whereas results that a paper reports under the name of a benchmark count even if it does not state which split it uses.
While reading, the two authors also record by name every benchmark that each included paper uses, adding benchmark names as they encounter them (open coding), and resolve disagreements in the same way.
We then track every benchmark that at least three of the 92 papers use, which yields 12 benchmarks, including the five benchmarks of~\autoref{table:benchmark_audit}.

\mypara{Counts}
\autoref{table:census_counts} lists the counts for all tracked benchmarks.
In this table and in~\autoref{table:benchmark_audit}, Only This counts the papers that use the benchmark but no other tracked benchmark.
Overall, 56 of the 92 papers report results on at least one tracked benchmark, 16 on exactly one and 40 on two or more.
Although \IMG is the most frequently used benchmark, it is the only tracked benchmark of just 6 of the 42 papers that use it.
Of these 42 papers, 29 also use at least one of \OSV, \YFCC, YFCC26k, and MP16-Reason-Test.

\begin{table}[t]
\centering
\caption{Number of the 92 census papers that report results on each tracked benchmark.}
\label{table:census_counts}
\scalebox{0.85}{
\begin{tabular}{@{}lrrr@{}}
\toprule
Benchmark & Papers & Only This & With Others \\
\midrule
\IMG~\cite{VJH17} & 42 & 6 & 36 \\
\YFCC~\cite{VJH17} & 23 & 0 & 23 \\
Im2GPS~\cite{HE08} & 15 & 2 & 13 \\
\OSV~\cite{ADSABFLNRVXZL24} & 13 & 2 & 11 \\
GWS15k~\cite{CKKCS23} & 7 & 0 & 7 \\
YFCC26k~\cite{MPE18} & 5 & 0 & 5 \\
IMAGEO-Bench~\cite{LYHLDZJ25} & 4 & 1 & 3 \\
DoxBench~\cite{LZLLZXX25} & 3 & 2 & 1 \\
MP16-Reason-Test~\cite{LZLTW25} & 3 & 0 & 3 \\
GeoBench~\cite{WLWHLR25} & 3 & 1 & 2 \\
Google Street View~\cite{JNHH25} & 3 & 1 & 2 \\
CityGuessr68k~\cite{KNS24} & 3 & 1 & 2 \\
\bottomrule
\end{tabular}
}
\end{table}

\section{OSV-5M Re-Split Details}
\label{appendix:split_details}

\mypara{Construction}
The pool of~\autoref{algorithm:split} leaves out the 129 images without a photographer ID.
Cells are the 11,399 cells of the official \OSV quadtree, and the 1,773 official test images outside all cells form their own bucket.
\autoref{table:split_parameters} lists the parameters and photographer counts of the re-split.
Specifically, $g_u$ counts the images of photographer $u$ in each (country, cell) pair only up to the remaining pair cap, which limits the score but not the later allocation.
Ties in the score and the image order within a pair follow SplitMix64 hashes of photographer and image IDs.
For the validation split, we skip photographers whose selection would leave a country below 90\% of its training images after the test split, which preserves the training data of small countries.
Finally, the 1 km buffer removes 128,751 training images rather than test images, so the test quotas stay fixed.

\begin{table}[t]
\centering
\caption{Parameters and photographer counts of~\autoref{algorithm:split}, where $t_p$ is the number of official test images in (country, cell) pair $p$.}
\label{table:split_parameters}
\begin{tabular}{@{}lrr@{}}
\toprule
 & Test Split & Validation Split \\
\midrule
Country Quota & $Q_k$ & $\mathrm{round}(0.02\,Q_k)$ \\
Pair Cap in the Score & $t_p$ & $\lceil 0.02\,t_p \rceil$ \\
Stopping Threshold $\lambda$ & 0.02 & 0.02 \\
Country Floor & None & 90\% \\
\midrule
Total Quota & 210,122 & 4,197 \\
Selected Photographers & 7,721 & 2,113 \\
Skipped by the Floor & 0 & 52 \\
Returned Photographers & 4,761 & 567 \\
Final Photographers & 2,960 & 1,546 \\
Final Images & 209,386 & 3,924 \\
\bottomrule
\end{tabular}
\end{table}

\mypara{Resulting Splits}
The test split reaches its country quota in 203 of the 219 countries, whereas the other 16 together fall short by 736 images.
Of its 209,386 images, 128,269 from 1,602 photographers are official test images and 81,117 are official training images.
The total variation distance over cells to the official test set is 5.36\%, mostly because 3,594 of the 9,805 pairs with official test images receive extra images when other cells of their country cannot be filled.
In addition, 15.83\% of the test images lie in cells without training images.

\mypara{Evaluated Test Images}
We read the inputs of PLONK, the QuadTree classifier, and nearest-neighbor retrieval from the WebDataset version of \OSV, which lacks 2,094 of the 210,122 official test images.
Since 1,297 of them belong to our test split, we evaluate on the remaining 126,972 of its 128,269 official test images, from 1,599 photographers.

\section{Implementation Details}
\label{appendix:implementation}

\mypara{Zero-Shot VLMs}
All four zero-shot VLMs receive the prompt in~\autoref{figure:zero_shot_prompt} from the released code of G3~\cite{JLLZWDHWWY24} and GeoRanker~\cite{JPGZL25}, followed by the test image.
The Instruct versions of Qwen2.5-VL-7B and Qwen2.5-VL-32B decode greedily with at most 128 new tokens.
In contrast, Gemini-3.8-Flash and GPT-6.1-Sol, which lack greedy decoding, use their default sampling and medium reasoning level with up to 40,000 and 16,384 output tokens, respectively.
Their APIs, queried on October 18, 2026 for \OSV and on November 4, 2026 for \YFCC, return the model names \texttt{gemini-3.8-flash} and \texttt{gpt-6.1-sol}.
We parse the first JSON object of each output or else, as G3 does, its first two decimal numbers.
In all cases, an output without valid coordinates, such as a refusal, counts as a failed prediction.

\begin{figure}[t]
\centering
\begin{tcblisting}{listing only, colback=gray!10!white, colframe=white, size=title, boxsep=1mm, listing options={basicstyle=\ttfamily\footnotesize, breaklines=true, breakatwhitespace=true, breakindent=1em, columns=fullflexible, keepspaces=true}}
Suppose you are an expert in geo-localization, you have the ability to give two number GPS coordination given an image.
Please give me the location of the given image.
Remember, you must have an answer, just output your best guess, don't answer me that you can't give a location.
Your answer should be in the following JSON format without any other information: {"latitude": float,"longitude": float}.
Your answer should be in the following JSON format without any other information: {"latitude": float,"longitude": float}.
\end{tcblisting}
\caption{Zero-shot prompt from the released code of G3 and GeoRanker. The code repeats the last line.}
\label{figure:zero_shot_prompt}
\end{figure}

\mypara{Retrained Methods}
We retrain PLONK~\cite{DPKL25} on each training set for 1,000,000 steps with its official code and \OSV configuration, which conditions on frozen StreetCLIP~\cite{HAS23} embeddings.
Likewise, we train the MLP of the QuadTree classifier~\cite{ADSABFLNRVXZL24} on frozen CLIP ViT-B/32~\cite{RKHRGASAMCKS21} features with the training settings of the \OSV paper, using QuadTree cells that we rebuild on each training set.

\mypara{Retrieval-Based Methods}
Nearest-neighbor retrieval uses exact cosine search over StreetCLIP features on \OSV and the top-ranked neighbor of G3 on \YFCC.
G3, GeoRanker, and GeoSearch keep their released weights.
Specifically, G3 embeds each \OSV training set with its released encoder, trained on \MPP, and searches each \YFCC database among its released \MPP embeddings.
Its variant with the open-weight VLM LLaVA-NeXT-8B generates coordinates from four prompts with 0, 5, 10, or 15 of the most and least similar database coordinates, as in its configurations for \IMG on \OSV and for \YFCC.
G3 then selects the candidate location whose location embedding best matches the image embedding.
GeoRanker, in turn, reranks the neighbors retrieved by G3 together with generated candidate locations, 12 retrieved and 3 generated on \OSV and 14 and 5 on \YFCC, as in its configurations for \IMG and \YFCC.
Its generated candidate locations come from LLaVA-NeXT-8B with the prompt in~\autoref{figure:zero_shot_prompt} on \OSV and from its released zero-shot list on \YFCC.
Finally, GeoSearch searches its released embeddings of the \OSV training set, restricted to each training set, with its encoder, also trained on \MPP.
We run its retrieval-only mode without web search, in which LLaVA-NeXT-8B outputs coordinates and the encoder selects one of them, and give LLaVA-NeXT-8B the query image, which this mode otherwise replaces with a blank image.
On \YFCC, its encoder predicts the most similar of 500,000 uniformly distributed GPS coordinates.

\mypara{GLOBE and GeoCLIP}
With its released weights and prompt, GLOBE~\cite{LZLTW25} outputs a city and a country, which we geocode with OpenStreetMap Nominatim instead of the Azure Maps service of its paper.
Unparsable or ungeocodable outputs, for 495 of the 5,000 \OSV images and 262 \YFCC images, are failed predictions.
GeoCLIP~\cite{CNS23}, by contrast, predicts the most similar of 100,000 \MPS GPS coordinates.

\mypara{Metrics}
On \OSV, the official evaluation code~\cite{ADSABFLNRVXZL24} measures the haversine distance $d$ in km between the prediction and the true location.
GeoScore averages $5000\exp(-d/1492.7)$ over test images, and Distance averages $d$ over valid predictions.
In addition, a country, region, or city prediction is correct if the reverse-geocoded names at this and all coarser levels match those of the test image.
On \YFCC, we instead compute geodesic distances, as G3 and GeoRanker do.
On both benchmarks, the accuracy at 1, 25, 200, 750, or 2,500 km is the share of test images with a distance error within that threshold.
Failed predictions count as misses under every accuracy metric and add zero to GeoScore.

\mypara{Memorization Check}
Since zero-shot VLMs could have seen the geotagged Flickr photos of \YFCC during pretraining, we count how often Gemini-3.8-Flash and GPT-6.1-Sol predict within 10 m or 100 m of the true location (\autoref{table:memorization}).
On \YFCC, only 0.55\% and 0.53\% of their predictions fall within 10 m, compared with 14.35\% for G3 with the original database, which contains 467 of the 4,536 test images.
Moreover, 8 of their 342 and 5 of their 260 predictions within 100 m match both true coordinates to four decimal places.
On \OSV, we keep 4,726 images of the 5,000-image subset after excluding the 274 whose burned-in text, transcribed with Qwen2.5-VL-7B, shows GPS coordinates that VLMs can read.
Taken together, both VLMs rarely reproduce the coordinates of test images.

\begin{table}[t]
\centering
\caption{Share (\%) of test images predicted within 10 m and 100 m of the true location.}
\label{table:memorization}
\scalebox{0.72}{
\begin{tabular}{@{}lcccc@{}}
\toprule
 & \multicolumn{2}{c}{\YFCC (4,536 Images)} & \multicolumn{2}{c}{\OSV (4,726 Images)} \\
\cmidrule(lr){2-3} \cmidrule(l){4-5}
Method & 10 m & 100 m & 10 m & 100 m \\
\midrule
Gemini-3.8-Flash & 0.55 & 7.54 & 0.21 & 0.44 \\
GPT-6.1-Sol & 0.53 & 5.73 & 0.15 & 0.32 \\
G3 (Original Database) & 14.35 & 17.42 & -- & -- \\
\bottomrule
\end{tabular}
}
\end{table}

\mypara{Tracing the Leakage Gain}
The analyses of~\autoref{section:mechanism} sort the distance between a database image and the query location into 16 bins with edges at 0.1, 0.2, 0.5, 1, 2, 5, 10, 20, 50, 100, 200, 500, 1,000, 2,000, and 5,000 km.
Specifically, the expected share in~\autoref{subsection:own_photographer} is, for each query, the share of its own photographer's images among all database images in the bin of its top-ranked neighbor, averaged over queries.
The replacement searches the top 128 neighbors on \OSV and the top 20 on \YFCC for the highest-ranked neighbor from another photographer.
The accuracies with the photographer-disjoint training set and the photographer-removed database that we compare with are computed on the same 115,461 and 2,598 queries.
In~\autoref{subsection:beyond_similarity}, two candidates have a matched distance if their distances $d$ to the query location (in km) fall into the same bin and their values of $\log_{10}(d + 0.1)$ differ by at most 0.05.
Such pairs exist for 52,660 queries from 890 photographers with nearest-neighbor retrieval and for 53,047 queries from 857 photographers with G3.
For the curves in~\autoref{figure:similarity_gap}, we group the pairs into bins of width 0.02 by similarity gap, centered at multiples of 0.02, and keep only the bins with at least 100 queries and 30 photographers.
Within each bin, we average over the pairs of each query before averaging over queries, so that every query has the same weight.
The preference gap at a similarity gap of zero is that of the bin centered at zero.
The similarity gap at which both images are preferred equally often is where a non-decreasing fit to these bins, weighted by their number of queries and linearly interpolated between bin centers, crosses zero.
Finally, for each retriever, the bootstrap of~\autoref{subsection:protocol} resamples the photographers of its queries with pairs.

\section{Complete Results}
\label{appendix:complete_results}

\autoref{table:complete_osv} reports country accuracy, GeoScore, and the accuracy at five distance thresholds for all 11 methods on the 5,000-image subset of \OSV, and~\autoref{table:complete_yfcc} reports the accuracy at the five thresholds for all 10 methods on \YFCC.
Furthermore,~\autoref{table:drops} lists the drop from the control to the photographer-disjoint reference data for each method whose reference data we swap.
Brackets give 95\% confidence intervals from the bootstrap over test photographers (\autoref{subsection:protocol}).
The results on the subset are close to those on 126,972 of the official test images in our test split.
With each of the three training sets, the country accuracy of PLONK, nearest-neighbor retrieval, and the QuadTree classifier differs by at most 0.90\% between the subset and these images.

\begin{table*}[t]
\centering
\caption{Results of all 11 methods on the 5,000-image subset of \OSV under each reference data condition. Brackets give 95\% confidence intervals, and shaded rows use photographer-disjoint reference data.}
\label{table:complete_osv}
\scalebox{0.70}{
\begin{tabular}{@{}llrrrrrrr@{}}
\toprule
 & & & & \multicolumn{5}{c}{Accuracy (\%)} \\
\cmidrule(l){5-9}
Method & Reference Data & Country Accuracy (\%) & GeoScore & 1 km & 25 km & 200 km & 750 km & 2,500 km \\
\midrule
PLONK & Official & 73.62 [69.71, 77.37] & 3,588.90 [3,476.23, 3,690.25] & 0.00 & 6.64 & 48.32 & 73.46 & 88.48 \\
 & Control & 73.74 [69.75, 77.58] & 3,591.57 [3,476.60, 3,693.59] & 0.00 & 7.08 & 49.16 & 73.46 & 87.92 \\
\rowcolor{gray!20}  & Photographer-Disjoint & 46.92 [40.80, 53.69] & 2,206.49 [2,092.16, 2,319.00] & 0.00 & 0.32 & 9.80 & 37.16 & 69.44 \\
GeoRanker & Official & 70.78 [66.64, 74.85] & 3,433.45 [3,289.11, 3,569.73] & 0.00 & 20.64 & 46.48 & 68.12 & 84.46 \\
 & Control & 70.14 [65.94, 74.30] & 3,406.10 [3,257.78, 3,544.27] & 0.00 & 20.12 & 46.38 & 67.18 & 84.28 \\
\rowcolor{gray!20}  & Photographer-Disjoint & 47.82 [42.25, 53.38] & 2,346.95 [2,219.19, 2,470.73] & 0.02 & 1.02 & 12.26 & 40.90 & 72.48 \\
Nearest-Neighbor Retrieval & Official & 70.74 [66.72, 74.78] & 3,395.15 [3,277.64, 3,502.14] & 0.00 & 21.26 & 46.58 & 66.22 & 84.34 \\
 & Control & 70.44 [66.31, 74.56] & 3,374.37 [3,255.25, 3,483.72] & 0.00 & 20.82 & 45.90 & 65.76 & 84.16 \\
\rowcolor{gray!20}  & Photographer-Disjoint & 42.38 [36.65, 48.95] & 2,046.90 [1,928.73, 2,162.88] & 0.00 & 0.58 & 8.00 & 33.62 & 65.74 \\
Gemini-3.8-Flash & -- & 66.42 [61.20, 71.75] & 3,283.97 [3,090.98, 3,506.35] & 6.96 & 15.66 & 37.20 & 64.50 & 84.58 \\
QuadTree Classifier & Official & 60.44 [55.74, 65.42] & 3,082.94 [2,943.03, 3,205.45] & 0.02 & 7.30 & 41.60 & 61.96 & 76.14 \\
 & Control & 58.84 [53.96, 63.89] & 2,995.95 [2,852.53, 3,123.47] & 0.04 & 6.28 & 39.30 & 60.26 & 74.64 \\
\rowcolor{gray!20}  & Photographer-Disjoint & 20.22 [17.53, 22.85] & 1,140.81 [1,048.48, 1,237.86] & 0.00 & 0.20 & 3.64 & 15.08 & 39.44 \\
GPT-6.1-Sol & -- & 57.90 [51.98, 63.99] & 2,834.94 [2,629.72, 3,093.67] & 6.08 & 11.70 & 26.80 & 53.42 & 77.46 \\
G3 & Official & 56.58 [51.12, 62.27] & 2,759.47 [2,620.97, 2,886.77] & 0.10 & 9.26 & 28.84 & 51.28 & 74.74 \\
 & Control & 55.44 [49.68, 61.38] & 2,718.77 [2,578.02, 2,852.36] & 0.10 & 8.38 & 28.04 & 50.54 & 74.18 \\
\rowcolor{gray!20}  & Photographer-Disjoint & 41.12 [34.86, 47.99] & 2,063.90 [1,925.88, 2,197.67] & 0.08 & 0.64 & 9.16 & 33.90 & 66.08 \\
GeoSearch & Official & 49.86 [44.93, 54.78] & 2,458.49 [2,317.36, 2,593.80] & 0.02 & 9.46 & 26.54 & 45.84 & 66.48 \\
 & Control & 49.18 [44.22, 54.18] & 2,432.37 [2,292.07, 2,567.87] & 0.06 & 8.92 & 25.46 & 45.08 & 66.10 \\
\rowcolor{gray!20}  & Photographer-Disjoint & 36.38 [30.67, 42.35] & 1,797.78 [1,670.00, 1,924.33] & 0.00 & 0.48 & 7.46 & 29.66 & 57.98 \\
GLOBE & -- & 44.82 [38.25, 51.85] & 2,066.80 [1,920.72, 2,206.74] & 0.14 & 1.78 & 11.14 & 35.24 & 64.82 \\
Qwen2.5-VL-7B & -- & 38.32 [31.41, 46.30] & 1,975.12 [1,697.56, 2,314.93] & 4.36 & 4.90 & 11.50 & 33.76 & 60.44 \\
Qwen2.5-VL-32B & -- & 34.88 [28.46, 42.58] & 1,645.95 [1,412.56, 1,910.51] & 2.34 & 3.24 & 8.94 & 27.32 & 51.16 \\
\bottomrule
\end{tabular}
}
\end{table*}

\begin{table*}[t]
\centering
\caption{Accuracy (\%) of all 10 methods on \YFCC under each retrieval database. Shaded rows use the photographer-removed database.}
\label{table:complete_yfcc}
\scalebox{0.85}{
\begin{tabular}{@{}llrrrrr@{}}
\toprule
Method & Retrieval Database & 1 km & 25 km & 200 km & 750 km & 2,500 km \\
\midrule
GeoRanker & Original & 32.65 & 43.41 & 53.97 & 69.82 & 82.32 \\
 & Control & 25.07 & 37.17 & 48.77 & 66.42 & 80.22 \\
\rowcolor{gray!20}  & Photographer-Removed & 12.13 & 24.36 & 38.21 & 59.57 & 77.29 \\
Nearest-Neighbor Retrieval & Original & 27.31 & 35.52 & 43.92 & 59.44 & 73.92 \\
 & Control & 18.87 & 28.33 & 37.59 & 54.78 & 71.05 \\
\rowcolor{gray!20}  & Photographer-Removed & 8.31 & 17.00 & 27.38 & 46.96 & 66.78 \\
Gemini-3.8-Flash & -- & 24.34 & 41.03 & 54.92 & 70.26 & 81.79 \\
G3 & Original & 24.03 & 33.91 & 44.07 & 61.09 & 75.60 \\
 & Control & 18.12 & 28.95 & 39.57 & 57.78 & 73.57 \\
\rowcolor{gray!20}  & Photographer-Removed & 9.08 & 18.85 & 30.75 & 51.15 & 70.19 \\
GPT-6.1-Sol & -- & 17.17 & 28.95 & 40.26 & 56.28 & 71.30 \\
GeoCLIP & -- & 9.59 & 19.29 & 32.63 & 54.98 & 74.67 \\
GLOBE & -- & 4.32 & 17.48 & 29.89 & 48.35 & 66.20 \\
Qwen2.5-VL-32B & -- & 2.95 & 13.23 & 24.45 & 43.43 & 59.61 \\
Qwen2.5-VL-7B & -- & 1.68 & 12.04 & 24.91 & 44.71 & 60.10 \\
GeoSearch Encoder & -- & 0.22 & 5.14 & 20.48 & 45.81 & 64.57 \\
\bottomrule
\end{tabular}
}
\end{table*}

\begin{table*}[t]
\centering
\caption{Drop from the control to the photographer-disjoint reference data, with 95\% confidence intervals, on the 5,000-image subset of \OSV and on \YFCC.}
\label{table:drops}
\scalebox{0.82}{
\begin{tabular}{@{}lrr@{\hskip 2em}lr@{}}
\toprule
\multicolumn{3}{c}{\OSV} & \multicolumn{2}{c}{\YFCC} \\
\cmidrule(r){1-3} \cmidrule(l){4-5}
Method & Country Accuracy (\%) & GeoScore & Method & 1 km Accuracy (\%) \\
\midrule
QuadTree Classifier & 38.62 [34.61, 42.45] & 1,855.15 [1,694.77, 1,999.56] & GeoRanker & 12.94 [11.65, 14.30] \\
Nearest-Neighbor Retrieval & 28.06 [24.53, 31.32] & 1,327.46 [1,216.31, 1,426.09] & Nearest-Neighbor Retrieval & 10.56 [9.45, 11.70] \\
PLONK & 26.82 [22.88, 30.85] & 1,385.08 [1,269.56, 1,493.93] & G3 & 9.04 [7.94, 10.16] \\
GeoRanker & 22.32 [19.83, 25.03] & 1,059.15 [938.10, 1,182.79] & \multicolumn{2}{c}{} \\
G3 & 14.32 [12.24, 16.48] & 654.87 [575.00, 732.93] & \multicolumn{2}{c}{} \\
GeoSearch & 12.80 [10.82, 14.81] & 634.59 [561.89, 701.13] & \multicolumn{2}{c}{} \\
\bottomrule
\end{tabular}
}
\end{table*}

\end{document}